# Differential Euler: Designing a Neural Network approximator to solve the Chaotic Three Body Problem

**Pratyush Kumar**
Department of Computer Science
Ashoka University
Sonipat, Haryana 131029
`pratyush.kumar_ug21@ashoka.edu.in`

**Aishwarya Das**
Department of Physics
Ashoka University
Sonipat, Haryana 131029
`anshu.das_ug21@ashoka.edu.in`

**Debayan Gupta**
Department of Computer Science
Ashoka University
Sonipat, Haryana 131029
`debayan.gupta@ashoka.edu.in`

## Abstract

The three body problem is a special case of the $n$-body problem where one takes the initial positions and velocities of three point masses and attempts to predict their motion over time according to Newton's laws of motion and universal gravitation.

Though analytical solutions have been found for special cases, the general problem remains unsolved (the solutions that do exist are impractical [1]). Fortunately, for many applications, we may not need to solve the problem completely [6] – *i.e.*, predicting with reasonable accuracy for some time steps, may be sufficient.

Recently, *Breen et al* [4] attempted to approximately solve the three body problem using (what appears to be) a simple neural network (NN). Although their methods appear to achieve some success in reducing the computational overhead, their model is extremely restricted, applying to a specialised two-dimensional case [13]. The authors do not provide explanations for critical decisions taken in their experimental design, no details on their model or architecture, and nor do they publish their code. Moreover, the model does not generalize well to unseen cases (where this new method might be an alternative to an expensive numerical integrator).

In this paper, we propose a detailed experimental setup to determine the feasibility of using neural networks to solve the three body problem up to a certain number of time steps. We establish a benchmark on the dataset size and set an accuracy threshold to measure the viability of our results for practical applications. Then, we build our models according to the listed class of NNs using a dataset generated from standard numerical integrators. We gradually increase the complexity of our data set to determine whether NNs can learn a representation of the chaotic three body problem well enough to replace numerical integrators in real-life scenarios.

## 1 Introduction

The 3-body problem has fascinated scientists for centuries and remains the oldest unsolved problem in classical mechanics. Despite everyone from Euler to Poincaré having tried their hand at it, there are only a handful of solutions for special cases of the 3-body problem known to us [14].

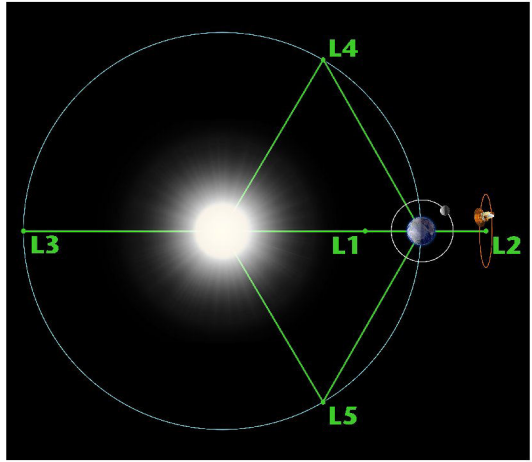

Figure 1: Lagrange Points in the Sun-Earth system:[2]

In the 1800s, Henri Poincaré proved that no general analytic solution exists for the general case of the three-body problem (though this does not rule out partial solutions). Euler found a family of solutions for three bodies orbiting around a mutual center of mass while remaining in a straight line (essentially a permanent ellipse in 3D space). Lagrange found solutions in which the three bodies form an equilateral triangle. [1]

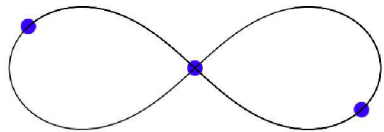

Figure 2: Figure 8 periodic solution

After Euler and Lagrange, there was a huge gap until the 1970's when Henon (Figure 3) and Broucke [5] found a family of solutions involving two masses bouncing back and forth in the center of a third body's orbit. In the 90's Cris Moore [19] discovered a stable figure 8-orbit [Figure 2] of three equal masses. The numerical discovery of the figure 8 solution was proved mathematically by Chenciner and Richard Montgomery [7] and the insights gained from their proof led to a discovery of large number of periodic three body orbits.

Recently in 2019, Nicholas Stone and Nathan Leigh [21] figured out that the motion of the three-body problem is perfectly deterministic when defined between two instances and can be thought of as a pseudo-random problem. Almost all three-body systems eject one of the bodies leaving a stable two-body system; the authors found that it was possible to identify regions of phase space where these ejections were likely to happen and we could then map out the orbital properties for the remaining two bodies.

To solve the three body problem, we break it down into tiny enough paths or time-steps. Then the small motions of all the bodies are updated step by step. This method is called numerical integration and has been used to accurately predict the motion of planets. Today, computationally expensive numerical integrators like Hermite [17], Brutus [3], etc. are used to calculate the solution to a certain level of precision for a fixed number of time steps in the future. Brutus is one of the few numerical integrators with high precision which makes it capable of computing converged solutions to almost any n-body problem. However, due to its high computational complexity ($\mathbf{O}(n^4)$), Brutus ends up being impractical for real world scenarios.

---

[1]If there are two bodies orbiting each other, the Euler and Lagrange solutions define 5 additional orbits for a third body that can be described analytically; these are the only perfectly analytical solutions to the three body problem. These spots are of special interest to us because if we place a low mass object in any of these 5 orbits, it will stay there indefinitely tracking the Earth's motion around the sun. These are called the Lagrange points and are used as spots to park our spacecraft.

[2]https://solarsystem.nasa.gov/resources/754/what-is-a-lagrange-point/

[3]https://rreusser.github.io/periodic-three-body-orbits/

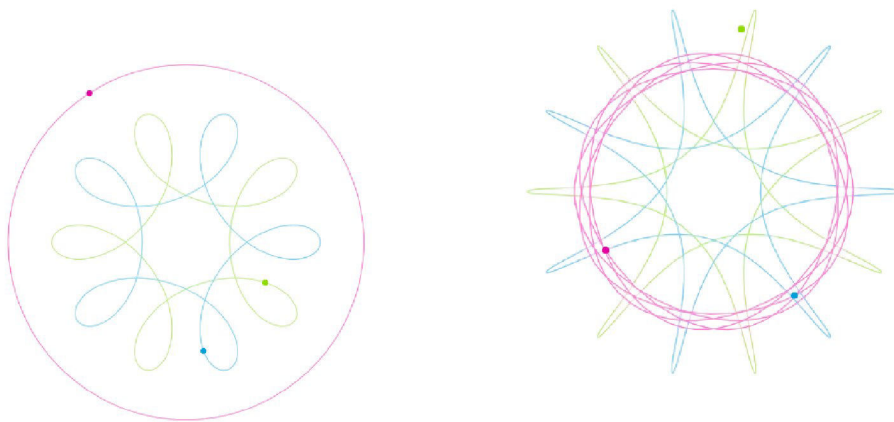

Figure 3: Broucke-H4 and Broucke-A16 orbits:[3]

Deep learning algorithms in the past few years have been very successful in simplifying complex tasks like image recognition, natural language processing, playing games, etc. by learning representations that generalize well. In this paper, we propose an experimental setup to find out how well NNs can solve the three-body problem. Specifically:

- Is it possible to learn a general representation of the three body problem?
- In terms of accuracy, are solutions to the three body problem given by NNs (see section 3.4) at par with those provided by numerical integrators?

## 2 Related Work

There has been a surge in the use of Neural Networks to learn chaotic behaviour exhibited by physical states and dynamical systems. *Breen et al* [4] train a Neural Network to learn the 3-body problem from data of the initial condition sampled from a 2D unit circle and future positions calculated using the Brutus simulator. They report good accuracy in the $t \in [0, 3.9]$ range but do not provide a proper benchmark; moreover, their model's accuracy decreases as the time frame of test data increases.

*Choudhary et al* [8] use Hamiltonian Neural Networks [10] to predict phase space trajectories in a chaotic nonlinear system. They train their network on multiple Hénon-Helies trajectories generated by numerically integrating Hamiltonian Equations. Their model is able to generalize and match the trajectories generated by numerical integration with very high accuracy.

*Pathak et al* [20] use Reservoir Computing methods, a class of networks derived from Recurrent Neural Networks that map input signals into higher dimensional spaces, to learn the Kuramoto-Sivashinsky equation to predict the evolution of a flame front. The resulting model was able to accurately predict the result for eight Lyapunov time [2].

We draw upon methods used in above work to design our experiments and test if a representation of the 3-body problem can be learned.

## 3 Methodology

In order to assess the ability of NNs to predict the trajectories of bodies in a three body system, we measure the capability of our models to predict the trajectories of the system past the time step t, up to which the model is trained upon. Additionally, we intend to test if the model is able to generalize and predict the trajectory of unseen systems(on which the model was not trained upon). Multiple iterations of the experiment will be conducted on:

- Different versions and representations of the problem, including:
  - Reduced 2D versions of the three body problem [13]
  - Full form of the three body problem (in 3D)
- With different network architectures:
  - Hamiltonian Neural Networks (HNNs)
  - Reservoir Computing models

### 3.1 Experimental Design

We hypothesize that certain neural networks like HNNs and reservoir computing systems will be able to predict the trajectory of the three bodies beyond 3 time steps in the future (for our reasoning, see section 3.2). We divide our tests into three levels of incremental complexity. Our model:

- Completely fails to predict the trajectory of the bodies for fewer than 3 time steps.
- Accurately predicts the trajectories for $T$ time steps into the future where $T \in [3,10]$.
- Accurately predicts the trajectories for $T$ time steps into the future where $T \in [10,100]$
- Accurately predicts the trajectories for $T$ time steps in the future where $T > 100$.

While testing, we will observe the Lyapunov time for which our model stays true to the actual trajectory and focus on investigating the area where we stop being able to predict well. This will allow us to measure the amount of power needed to improve the prediction capabilities beyond a certain time step. We have chosen 10s and 100s to determine if order-of-magnitude improvements to current results can be achieved using our methods. *Breen et al* [4] have used 10k samples to predict trajectories upto 3s with high accuracy, so we use this as a starting point for comparison. We will test each of our hypothesis by calculating the average error for each trajectory and then calculate our level of confidence for 90% , 95 % and 98 % respectively.

### 3.2 Network Architectures

We choose two structurally different classes of Neural Networks to run our test. First is the physics inspired Hamiltonian Neural Networks.

Hamiltonian Neural Networks (HNNs) [10] work on the principle of the total energy of a dynamic system being conserved. Using Hamiltonian mechanics, the total energy of a system can be related to the position (**q**) and momentum (**p**) of a system as U = $\mathcal{H}$(**q**,**p**). Since the Symplectic gradients of a system's Hamiltonian – $\frac{\partial \mathcal{H}}{\partial \mathbf{p}}$ and $\frac{-\partial \mathcal{H}}{\partial \mathbf{q}}$ – tell us the time evolution of a system with position **q** and momentum **p**, HNNs, given **q** and **p**, try to learn these gradients in order to learn the representation of the given system. During the forward pass, HNNs learn a scalar quantity ($\mathcal{H}_\theta$) analogous to the total energy of a system. The gradient of this quantity is calculated with respect to **p,q** and the loss is calculated as shown in 1 below.

$$\mathcal{L}_{HNN} = \left\| \frac{\partial \mathcal{H}_\theta}{\partial \mathbf{p}} - \frac{\partial \mathbf{q}}{\partial t} \right\|_2 + \left\| \frac{\partial \mathcal{H}_\theta}{\partial \mathbf{q}} + \frac{\partial \mathbf{p}}{\partial t} \right\|_2 \tag{1}$$

The three body problem is inherently a chaotic problem which means it's very sensitive to the initial conditions of the system. A conventional NN setup will predict the state of the system given current state. The problem here stems from the fact that while time is actually continuous, the data given to NNs is actually discrete. Since the system is chaotic, the discrete data is unable to capture a complete picture of evolving states of the system. This introduces small errors in the calculation as the training progresses. Because of the error associated with each computation, training the data using a Neural network has not given good results except for small time steps as numerical noises lead to large deviations of numerical solutions for the three body problem. We plan to apply HNNs to a three body system to correctly predict their orbits. This decision is fueled from the fact that for solving such problems, physicists often use the invariant quantities to predict the dynamics of their system. These invariant quantities could be the total energy or the total momentum of the system. The Hamiltonian approach is a powerful way of solving any problem where total energy of a system is conserved. Moreover, the solution of the three body problem exists in the form of a differential equation and it has been shown that Hamiltonian Neural Networks are capable of solving differential equations [18] which makes Hamiltonian Neural Nets a viable option.

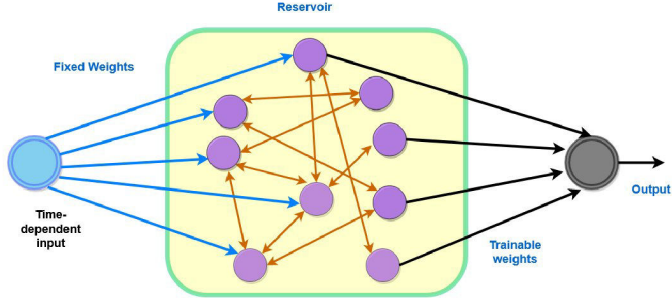

Figure 4: Structure of a Reservoir Computing setup

Another model we include in our experimental setup is Reservoir Computing methods [22], a derivative of Recurrent Neural Networks capable of mapping inputs into higher dimensional spaces. The general structure of a reservoir computing setup is depicted in figure 4 [9]. Inputs are connected to the reservoir (a collection of random number of nodes randomly connected) via neurons whose weights are untrainable and an output is given via regression over the reservoir nodes.

Reservoir computing methods, although similar to RNNs, make a conceptual and computational separation between **h(n)** – a dynamic *reservoir*, an RNN as a nonlinear temporal expansion function, and **y(n)** – a recurrence free *readout* that produces the desired output from the expansion [15]. The separation is based on the understanding that the recurrent states **h(n)** and the output **y(n)** serve different purposes. **h(n)** expands the input history **x(n), x(n-1)**, ... into a rich enough $R^{N_x}$ dimensional reservoir state space while **y(n)** combines the neuron signals **h(n)** into a desired output. Due to this conceptual difference, both these sections are trained separately. The only weights trained are the $\mathbf{W_{out}}$ responsible for generating a suitable linear combination of the reservoir nodes. A reservoir is set up as

$$\mathbf{h(n)} = f(\mathbf{W}\mathbf{h_{(n-1)}} + \mathbf{W_{in}}\mathbf{x(n)})$$

where f is the the non-linearity function (eg. tanh), $\mathbf{h_{(n-1)}}$ the previous reservoir state, **x(n)** the input for current time step and **W** and $\mathbf{W_{in}}$ the respective weights. The readout is a simple linear combination of the nodes in a reservoir and is set up as

$$\mathbf{y_n} = f(\mathbf{W_{out}}\mathbf{h(n)})$$

While there exist multiple methods to set up reservoir computing, the two most often used are a) Echo State Networks (ESN) [12] and b) Liquid State Machines (LSM) [16]. The only major conceptual difference between ESN and LSM is that the later uses Spiking neural networks instead of feed-forward neural networks which results in each neuron receiving time varying inputs resulting in a spatio-temporal pattern of activation in the network.

The untrained reservoir is expected to have a rich set of dynamics so that the effect of previous inputs and states vanishes gradually as time passes. This property, in the case of ESNs, is called the *echo state property*. The weight matrix **W** for the reservoir, generally, are large, sparse and randomly generated from a uniform distribution symmetric around zero while the input weights $\mathbf{W_{in}}$ are dense. For most practical purposes, the echo state property is assured if the reservoir weight matrix **W** is scaled so that its spectral radius $\rho(\mathbf{W})$ (the largest absolute eigenvalue satisfies) $\rho(\mathbf{W}) < 1$ [15]. Jaeger et al. [12] also suggest that $\rho(\mathbf{W})$ should be close to 1 for tasks that require long memory and smaller for tasks where retaining a longer memory could be detrimental.

Reservoir Computing methods have been successful in predicting time-series data for chaotic spatio-temporal dynamical systems as shown in [9] along with other cases discussed in the previous section. Since, the 3-body problem is a time-series, this makes Reservoir computing a viable approach.

### 3.3 Problem versions and dataset generation

Following in the footsteps of previous efforts [4, 14], we start with a two dimensional representation of the problem. The data will be of the form *[(x,y),t]* for each of the three bodies. Similarly, for the three dimensional case, the data format will be *[(x,y),z,t]* for each of the three bodies. Data is generated using Brutus for both the cases. A total of *10000* samples are generated for the training set and *500* samples for the test set where all the samples in both the training and the test set consist of *100* time steps.

### 3.4 Training

Since the three body problem is inherently a time-series problem, we will use an LSTM [11] to establish a bottomline benchmark on the predictive capability of Neural Networks. The LSTM is trained on both 2D and 3D versions of the problem, and outputs the position of the objects in future time steps.

Moving ahead, we will train both Hamiltonian Neural Network and Reservoir Computing models on the periodic case for the 2D representation of the three body problem. First we will train on the general case of the problem in 2D. Results will be noted as mentioned in the Experimental Setup above. The same steps as above will be repeated for the dataset recorded in 3D and the results will be tested accordingly.

## 4 Conclusion

We use two different classes of neural networks and see how well they approximate the three body problem. Vanilla neural networks haven't been successful in approximating chaotic problems, but by using physics inspired neural networks we hope to learn the underlying order and chaos of the three body problem.